\def\year{2020}\relax
\documentclass[letterpaper]{article} 
\usepackage{latexsym,amsfonts,amsmath,diagbox,lipsum,booktabs,siunitx,float,multirow}
\usepackage{aaai20}  
\usepackage{times}  
\usepackage{helvet} 
\usepackage{courier}  
\usepackage[hyphens]{url}  
\usepackage{graphicx} 
\usepackage{graphicx}  
\title{On the Generation of Medical Question-Answer Pairs}
\author{
 Sheng Shen\textsuperscript{\rm 1},
 Yaliang Li\textsuperscript{\rm 2},
 Nan Du\textsuperscript{\rm 3},
 Xian Wu\textsuperscript{\rm 3}, \\
 \bf \Large
 Yusheng Xie\textsuperscript{\rm 3},
 Shen Ge\textsuperscript{\rm 3},
 Tao Yang\textsuperscript{\rm 3},
 Kai Wang\textsuperscript{\rm 3},
 Xingzheng Liang\textsuperscript{\rm 3},
 Wei Fan\textsuperscript{\rm 3}
 \\
\textsuperscript{\rm 1}University of California at Berkeley,
\textsuperscript{\rm 2}Alibaba Group,
\textsuperscript{\rm 3}Tencent\\
sheng.s@berkeley.edu, yaliang.li@alibaba-inc.com, \\ \{ndu, kevinxwu, yushengxie, shenge, tytaoyang, ironswang, evelynliang, Davidwfan\}@tencent.com
}

\begin{document}
\maketitle

\begin{abstract}
Question answering (QA) has achieved promising progress recently. However, answering a question in real-world scenarios like the medical domain is still challenging, due to the requirement of external knowledge and the insufficient quantity of high-quality training data. In the light of these challenges, we study the task of generating medical QA pairs in this paper. With the insight that each medical question can be considered as a sample from the latent distribution of questions given answers, we propose an automated medical QA pair generation framework, consisting of an unsupervised key phrase detector that explores unstructured material for validity, and a generator that involves a multi-pass decoder to integrate structural knowledge for diversity. A series of experiments have been conducted on a real-world dataset collected from the National Medical Licensing Examination of China. Both automatic evaluation and human annotation demonstrate the effectiveness of the proposed method. Further investigation shows that, by incorporating the generated QA pairs for training, significant improvement in terms of accuracy can be achieved for the examination QA system. \footnote{Our full version paper with supplemented material is publicly available at \url{https://arxiv.org/abs/1811.00681}.}
\end{abstract}

\section{Introduction}

Due to the remarkable breakthrough of deep learning and natural language processing, question answering (QA) has gained increasing popularity in the past few years. Among QA's broad application domains, medical QA is one of the most appealing real-world application scenarios: People tend to consult others about health-related issues on online communities, which might be more affordable than visiting doctors in resource-limited areas.

Although QA systems with deep learning methods have achieved good performance, medical QA confronts particular difficulties against other domains. First, medical QA system requires highly accurate answers, and thus external and professional knowledge gathered from various sources are needed. Second, the size of available high-quality medical QA pairs is limited, as the labeling process by medical experts is time-consuming and expensive. Therefore, the performance of medical QA system is further constrained by the paucity of high-quality QA pairs since it can hardly learn a good model from limited training data. Though~\cite{roberts2017overview,pampari2018emrqa} aim to enrich the dataset itself, but the efforts are still far from enough.

To tackle these difficulties, the generation of medical QA pairs plays an indispensable role. By automatic generation of high-quality medical QA pairs, external and professional knowledge can be incorporated, and the size of training data can be augmented. Therefore, we study this important task of medical QA pair generation in this paper. To be more specific, we assume that each medical answer corresponds to a distribution of valid questions, which should be constrained on external medical knowledge. Following this assumption, with more high-quality QA pairs generated based on the same knowledge as original QA pairs, the latent distribution of available medical QA pairs can be supplemented and thus medical QA system could learn unbiased model easily.

However, the generation of new medical QA pairs based on original ones is challenging: It is hard to simultaneously maintain the diversity and the validity of generated question-answer pairs. Existing question-answer pair generation methods \cite{yang2017semi,song2018leveraging} either has external context to build upon or  \cite{duan2017question,CardieD18,yang2017semi} focused more on the word-level similarity, and it may generate lexically similar question-answer pairs to the original ones. These generated similar QA pairs are valid but of limited use for allowing the system to answer questions involving new knowledge. On the other hand, if more diversity in the discourse/sentence level is promoted, validity might not be guaranteed.

\begin{figure*}
\begin{center}
\includegraphics[width=0.8\linewidth]{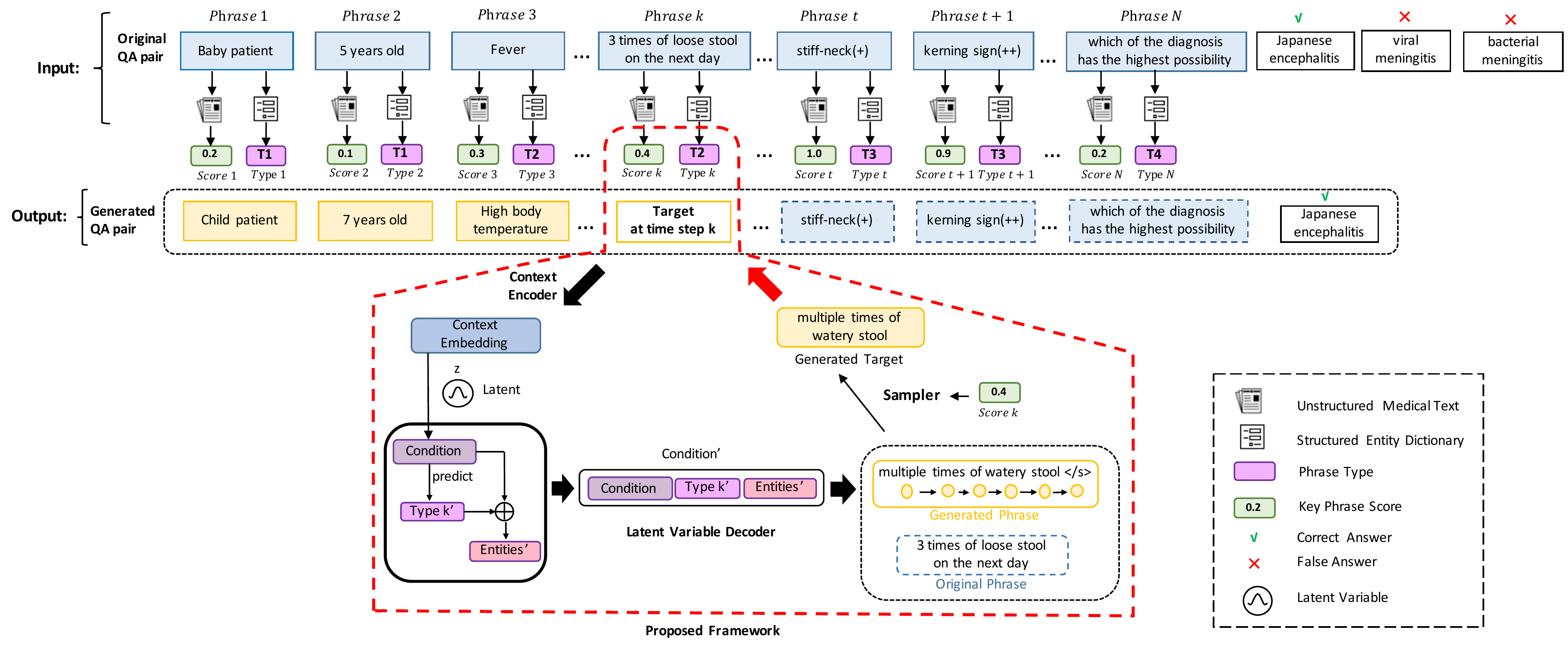}
\caption{Overview of the proposed framework. Note that this question consists of $N$ phrases and this figure shows the process where we are generating the $k$-$th$ phrase.}\label{fig:overview}
\end{center}
\end{figure*}

To ensure the validity of the generated medical QA pairs, we propose a retrieval and matching method to detect the key information of QA pairs in an unsupervised way using unstructured text materials such as patients' medical records, textbooks, and research articles.

To promote the diversity of the generated medical QA pairs while retaining validity, we propose two mechanisms to incorporate structured, unstructured knowledge for QA generation. We first explore global phrase level diversity and validity with a hierarchical Conditional Variational Autoencoder (CVAE) framework, which models phrase level relationship in original medical QA pairs, and generates the new pairs without breaking these relationships. We then propose a multi-pass decoder, in which all the local components (phrase type, entities in each phrase) are coupled together and are jointly optimized in an end-to-end fashion.

In order to demonstrate the effectiveness of the proposed generation method, we evaluate generated medical QA pairs through qualitative and quantitative measures, and the results confirm the high-quality of the generated medical QA pairs. Further, in an application of the proposed method to a medical certification exam, the experimental results show that the generated medical QA pairs improve the original QA system by six percent question-level accuracy. 

\section{Methodology}

In this section, we introduce our framework for generating medical question-answer pairs based on existing pairs. For medical QA, we assume the same answer can be produced by multiple questions, for example, patients of \emph{stiff neck}(+) with \emph{pap test}(+) or \emph{respiratory failure} can be diagnosed as the disease \emph{Japanese encephalitis} due to the diversity of medical characteristics, while for a specific medical question, there is only one correct answer. Hence, we view the generating process of medical QA pairs as generating questions given a certain answer. Technically speaking, our framework for generating medical QA pairs can be considered as an approximation of the latent distribution of questions given answers and sampling new questions from the distribution. As shown in Fig~\ref{fig:overview}, the whole framework involves a key phrase detector and an entity-guided CVAE based generator (eg-CVAE), which we  describe in detail in the following subsections.
Both the original QA pairs and the generated ones from our framework will be fed into the QA system as inputs for training.

\begin{figure*}
\begin{center}
\includegraphics[width=0.8\textwidth]{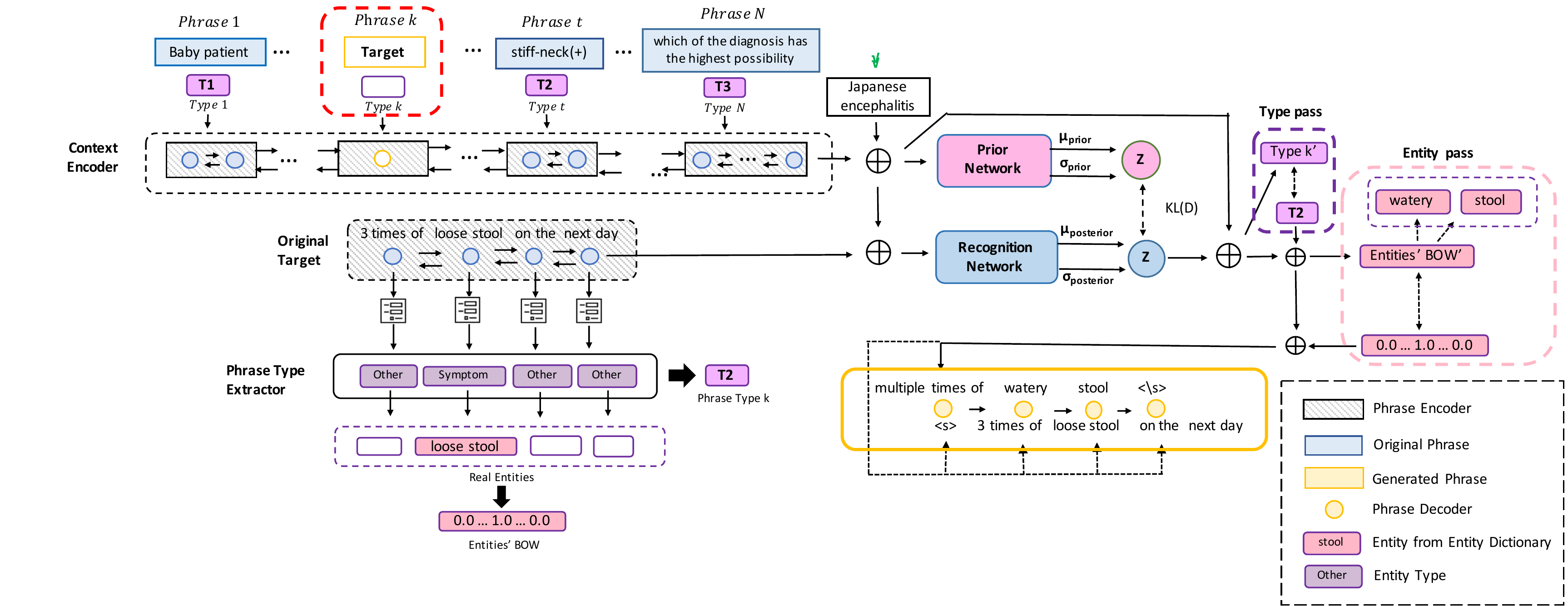}
\caption{Entity-guided CVAE based Generator. In this figure, we illustrate the detailed process to generate current phrase$_k$ based on previous altered phrases$_{1,...,k-1}$, $_{k+1, N}$.}\label{fig:generator}
\end{center}
\end{figure*}

\subsection{Key Phrase Detector}

In order to approximate the unknown conditional distribution of medical questions given answer, we leverage external knowledge to exploit the intrinsic characteristics of medical questions that associate with the same answer. Specifically, every medical question $Q$ consists of several phrases $P_k, k\in[1, N]$, such as patient's symptoms, examination results. Each phrase is composed of several words. Among medical questions, there exist \emph{key phrases} highly correlated with the answers (denoted as $P_k'$ like \emph{stiff neck}(+) in Fig~\ref{fig:overview}). To detect the prior key phrases, we employ an unsupervised matching approach on unstructured medical text. Furthermore, to ensure the consistency of these key phrases in the generated new questions, we assign each phrase with a normalized significance score $s_k \in[0,1]$, which is further used as the probability of replacing this phrase by the generated one or not in the generation process.

Rather than considering each phrase separately, we assume that the co-occurrence probability of a key phrase and answer indicates the significance of that phrase. To explore this co-occurrence information, we first use each medical QA pair as query to perform an Elasticsearch\footnote{https://github.com/elastic/kibana} \cite{gormley2015elasticsearch} based retrieval over the medical materials. We also apply rules to ensure the presence of the answer in retrieved texts, denoted as $R_i, i\in[1, M]$ (M stands for the number of retrieved texts). An unsupervised matching strategy is proposed to model the relevance of a certain phrase $P_k$ with the answer by matching $P_k$ with all the $R_i$. Specifically, we divide each $R_i$ into phrases $P^{R_i}$ (each phrase contains multiple words), and represent each $P^{R_i}$ and $P_k$ into the same vector space. To produce that vector, we perform a hierarchical pooling over the word embedding $v_j, j\in[1, L]$ in that phrase following \cite{baselove}: first, average pooling over $v_{j, j+k-1}, j\in[1, L-k+1]$ within each sliding window (size is $k$); then, max pooling over the induced average-pooling vectors. We match every phrase $P_k, k\in[1, N]$ with the phrase splits $P^{R_i, i\in[1, M]}$ using cosine distance and store the highest score $s_k^{R_i}$. The unnormalized matching score for $P_k$ with $R$ is the mean value of $s_k^{R_i}, i\in[1,N]$. These scores for each phrase $P_k$ in the QA pair will be normalized as the $s_k, s_k\in[0,1]$ for final sampling decision with the Min-Max method. Specifically, in inference , we randomly samples $p^\prime_k\in[0,1]$. Then if $p^\prime_k>s_k$ we will replace $P_k$ with the generated phrase or retains $P_k$.

\subsection{Entity-guided CVAE based Generator} 
A medical question has two levels of structures: one structure exists within a single phrase, which is dominated by local information of involved medical entities, and the other is a distinct across-phrase structure, which is characterized by aspects such as phrase types and the corresponding answer etc.. We thus explore the answer conditioned medical question generation in a two-level hierarchy: sequences of sub-sequences (iterative phrase generation process), and sub-sequences of words. Towards modeling the constraint over the whole question, we first use Conditional Variational Autoencoder. Moreover, towards modeling the internal structure within each phrase, we draw the idea from human's process to generate a complete question (start from a sketch and then details), and introduce a three-pass decoding process: first implicit type modeling, then explicit entities modeling, and finally phrase decoding.

\subsubsection{Conditional Variational Autoencoder}
Motivated by ~\cite{serban2017hierarchical}, we adapt the original CVAE for dialog generation to our setting by considering question generation as an iterative phrase generation process in Figure \ref{fig:generator}. To this end, we represent each phrase generation procedure with three random variables: the phrase context $c$, the target phrase $x$, and a latent variable $z$ that is used to capture the latent distribution over all valid phrases. For each phrase, $c$ is composed of both the sequence of other phrases in the question and the corresponding answer. We then define the conditional distribution $P(x, z|c) = P(x|c, z) \cdot P(z|c)$ and set the learning target is to approximate $P(z|c)$ and $P(x|c, z)$ via deep neural networks (parametrized by $\theta$). We refer to $P_\theta(z|c)$ as the prior network and $P_\theta(x|c, z)$ as the target phrase decoder. Then the generative process of $x$ is summarized as first sampling a latent variable $z$ from $P_\theta(z|c)$ (a parametrized Gaussian distribution.) and then generating $x$ by $P_\theta(x|c, z)$.

The CVAE is trained to maximize the conditional log likelihood of $x$ given $c$, meanwhile minimizing the KL divergence between the posterior distribution $P(z|x, c)$ and a prior distribution $P(z|c)$. We assume that both $z$ follow multivariate Gaussian distribution with a diagonal covariance matrix. Further, we introduce a recognition network $Q_\phi(z|x, c)$ to approximate the true posterior distribution $P(z|x, c)$. As proposed in \cite{sohn2015learning}, CVAE can be efficiently trained with the \textit{Stochastic Gradient Variational Bayes} (SGVB) framework \cite{kingma2013auto} by maximizing the variational lower bound of the conditional log likelihood, which can be written as:
\begin{equation}
\begin{aligned}
\mathcal{L}(\theta, \phi; x, c) &\ = -KL(Q_\phi(z|x, c)||P_\theta(z|c)) \\ 
&\ + E_{Q_\phi(z|x, c)}[\log P_\theta(x|c, z)].
\end{aligned}
\end{equation}

At timestamp $k$ of the whole generation process to produce a question phrase, the phrase encoder is a bidirectional recurrent neural network \cite{schuster1997bidirectional} with a gated recurrent unit (GRU \cite{chung2014empirical}) to encode each phrase $P_k$ into a fixed-size vector by concatenating the last hidden states of the forward and backward RNN as $[\mathop{hv_k}\limits ^{\rightarrow}, \mathop{hv_k}\limits ^{\leftarrow}]$. This basic phrase context encoder is a one-layer GRU network that encodes the $N-1$ context phrases (in training, the context phrases are from the original question; in testing, the preceding $k-1$ phrases are from the generated question) as $hv_{1:k-1}$ with $hv_{k+1:N}$. The last hidden state $hv^c$ of the phrase context encoder is concatenated with the corresponding answer embedding $a$ and $c = [hv^c, a]$. As we assume $z$ follows an isotropic Gaussian distribution, the recognition network $Q_\phi(z|x, c) \sim N(\mu, \sigma^2I)$, the prior network $P_\theta(z|c) \sim N(\mu', \sigma'^2I)$, and then we get:
\begin{align}
\begin{bmatrix} \mu \\ \log(\sigma^2) \end{bmatrix} = W_r  \begin{bmatrix} x \\ c \end{bmatrix} + b_r, 
\begin{bmatrix} \mu' \\ \log(\sigma'^2) \end{bmatrix} = MLP_p(c).
\end{align}

The reparameterization trick \cite{kingma2013auto} that uses formed parameter to treat $z$ as deterministic node is adopted to get samples from $N(z;\mu, \sigma^2I)$ in training (recognition network) and from $N(z;\mu', \sigma'^2I)$ in testing (prior network). The final phrase decoder at timestamp $k$ is a one-layer GRU network with initial state set as $W_k[z, c] + b_k$. The words will be predicted sequentially by the phrase decoder.

\subsubsection{Phrase-type Augmented Encoder}
Inspired by \cite{parvez2018building}'s insights to facilitate text generation with entity type, we similarly introduce phrase type in the medical domain as a similar source of structural information (the intuition behind specific phrases such as lab examination and physical characteristics employed by doctors). Rather than focusing on word level, we assume each phrase information involves two levels of characteristics: 1) global characteristic as the surrounding or context phrases' type information; 2) local characteristic as entity type knowledge within each phrase. Moreover, to address the difficulty of acquiring labeled data from experts, we propose to directly utilize a structured entity dictionary and model the phrase type in a contextualized way following~\cite{Elmo}.

To this end, we design a sequence labeling task for pre-training, whose learning goal is to predict each word's type (for those words not in the entity dictionary, the type is considered as ``other'') over the whole question. 

A Bi-LSTM-CRF model, which takes each word's embedding in the question as input and their types as output, is applied in the pre-training task. We use Bi-LSTM layer to encode word-level local features, and CRF layer to capture sentence-level type information. As the pre-training task's accuracy can achieve $97.08\%$, we assume that the hidden states of Bi-LSTM for each word $k$ as $h_k [\mathop{h_k}\limits ^{\rightarrow}, \mathop{h_k}\limits ^{\leftarrow}]$ can encode the contextualized type information. Considering that each phrase can be split into multiple words, the phrase type information is introduced by performing max-pooling over each word's $h_k$. We then concatenate contextualized type vector $t_k$ at timestamp $k$ to generate phrase type vector $hv_k' = [hv_k, t_k]$ for $P_k$ (clustering as 6 $T_*$ in Figure \ref{fig:generator}). $t_k$ is pre-trained through the sequence labeling task, and different for each timestamp of the whole generation procedure. The new $x' = hv_k'$ will be then applied for the recognition network.

\subsubsection{Entity-guided Decoder}
Other than only conditioning on the corresponding answer, we introduce extra constraints on latent $z$ to keep it meaningful during decoding process. Drawn the insights from the process of human generating a complete question (start from a sketch and then details) in \cite{xia2017deliberation}, we propose a multiple pass decoding procedure to incorporate inter-phrase level and intra-phrase level information as constraints. We thus model the contextualized type $t$, which is imposed by the entity dictionary, at the first pass to ensure the consistency of type information across phrases. We then conjecture entities to be the skeleton within each phrase, and explicitly model entities $e$ at the second pass. We promote diversity in our generation process by adding entity-level variation during inference, allowing the production of phrases with similar semantics towards the same answer but containing diverse entities.

We assume that the generation of phrase $P_k$ as $x$ depends on $c$, $z$, $t$ and $e$; $e$ relies on $c$, $z$, $t$; and $t$ relies on $c$, $z$. During training, the initial state of the final decoder is $d_k = W_k[z, c, t, e] + b_k$ and the input is $[w_{1:n^k}, t, e_k]$ where $w_{1:n^k}$ is the word embedding of words in $x$ and $e_k$ is average pooling embedding of the entire entity embedding in $x$. In the first type-prediction pass, there is an MLP to predict $t' = MLP_t(z, c)$ based on $z$ and $c$. In the second entity-prediction pass, another MLP is used to predict $e_{\rm softmax'} = MLP_e(z, c, t)$ based on $z$, $c$ and $t$. Then $e_{\rm softmax'}$ is multiplied with the whole entity embedding matrix for the aggregation of the $e_k'$. In the testing stage, the predicted $t'$ and $e_k'$ are used in the final phrase decoder.

\subsection{Training Objective}
To induce meaningful latent variable $z$, we explicitly model the generation of $x$ as a multi-pass process, which might relieve the posterior collapse problem \cite{he2019lagging} motivated by~\cite{zhao2017learning} in enriching the information in posterior distribution of $z$ with dialog actions.

Specifically, by introducing phrase-type information in the first pass, we suppose that the generation of $x$ is based on $c,z$ and $t$, where $t$ is based on $c$. Then the modified variational lower bound for eg-CVAE without entity modeling:
\begin{equation}
\begin{aligned}
\mathcal{L}(\theta, \phi; x, c, t) &\ = -KL(Q_\phi(z|x, c, t)||P_\theta(z|c)) \\
&\ + E_{Q_\phi(z|x, c, t)}[\log P_\theta(t|c, z)] \\
&\ + E_{Q_\phi(z|x, c, t)}[\log P_\theta(x|c, z, t)].
\end{aligned}
\end{equation}

To refine phrase-type information into detailed entities in the second pass, we model $e$ explicitly based on the assumption that the produce of $x$ is divided into two phases: exploiting phrase-type to generate $e$; and using $e,t,c$ and $z$ to generate $x$. Thus the final eg-CVAE model is to maximize: 
\begin{equation}
\begin{aligned}
\mathcal{L}(\theta, \phi; x, c, t, e) &\ = -KL(Q_\phi(z|x, c, t, e)||P_\theta(z|c)) \\ 
&\ + E_{Q_\phi(z|x, c, t, e)}[\log P_\theta(t|c, z)] \\
&\ + E_{Q_\phi(z|x, c, t, e)}[\log P_\theta(e|c, z, t)] \\
&\ + E_{Q_\phi(z|x, c, t, e)}[\log P_\theta(x|c, z, t, e)].
\end{aligned}
\end{equation}

Furthermore, the KL annealing~\cite{serban2016generating} technique as gradually increasing the weight of the KL term from $0$ to $1$ during training and auxiliary bag-of-words loss of $x$ as in~\cite{zhao2017learning} are also adopted.

\section{Experiments}
\subsection{Dataset}
To validate the effectiveness of the proposed method, we collect real-world medical QA pairs from the National Medical Licensing Examination of China (denoted as NMLEC\_QA). The collected NMLEC\_QA dataset contains $18,798$ QA pairs, and we generate new QA pairs based on these original ones. We adopt NMLEC\_2017 as the test set to evaluate the QA system, which will not be used in QA pair generation. The medical entity dictionary is extracted from medical Wikipedia-style pages\footnote{\url{http://www.xywy.com/}}, and the constructed dictionary covers $19$ types of medical entities. The unstructured medical materials consists of $2,130,128$ published paper in medical domain and $518$ professional medical textbooks. 

\begin{table*}
\centering
\small \caption{Performance comparison under automatic evaluation metrics.}
\label{auto_result}
\begin{small}
\begin{tabular}{ccccccccccc}
\toprule
\multirow{2}{*}{\textbf{Method}} & \multicolumn{3}{c}{\textbf{BLEU}} & \multicolumn{3}{c}{\textbf{BOW Embedding}} & \multicolumn{2}{c}{\textbf{intra-dist}} & \multicolumn{2}{c}{\textbf{inter-dist}} \\ 
\cmidrule{2-4} \cmidrule{5-7} \cmidrule{8-9} \cmidrule{10-11} 
 & Precision & Recall & F1 & Average & Extreme &Greedy & dist-1 & dist-2 & dist-1 & dist-2 \\ 
 \midrule
HRED & 0.435 & 0.737 & 0.547 & 0.753 & 0.705 & 0.809 & 0.837 & 0.912 & 0.205 & 0.255 \\
VHRED & 0.454 & 0.705 & 0.533 & 0.863 & 0.872 & 0.887 & 0.803 & 0.991 & 0.562 & 0.538 \\
type-CVAE & 0.507 & 0.748 & 0.572 & 0.872 & 0.852 & 0.892 & 0.831 & \textbf{0.997} & 0.555 & 0.581 \\
entity-CVAE & \textbf{0.541} & \textbf{0.781} & \textbf{0.613} &\textbf{0.891} & \textbf{0.903} & \textbf{0.874} & 0.840 & 0.996 & 0.533 & 0.554 \\ 
eg-CVAE & 0.450 & 0.611 & 0.494 & 0.802 & 0.793 & 0.819 & \textbf{0.867} & 0.994 & \textbf{0.637} & \textbf{0.589} \\ 
\bottomrule
\end{tabular}
\end{small}
\end{table*}

\subsection{Baselines}
We compare the performance of the proposed method \textbf{eg-CVAE} with two recently-proposed text generation methods: \emph{HRED}~\cite{SerbanSBCP16}, a sequence-to-sequence model with a hierarchical RNN encoder, and \emph{VHRED}~\cite{serban2017hierarchical}, a hierarchical conditional VAE model. We also test the contribution of the multiple steps of our decoder of type modeling or entity modeling process: \emph{type-CVAE} with type decoding as the only-pass, and \emph{entity-CVAE} with entity decoding as the only-pass.

\subsection{Evaluation based on Automatic Metric}
Automatically evaluating the quality of generated text remains challenging ~\cite{LiuLSNCP16}, and thus we design automatic evaluation metrics for our specific scenario.
As mentioned above, we assume that each QA pair can be considered as a question sampled from a latent answer-conditioned distribution. Based on each original question-answer pair, we generate $N$ new questions by iteratively sampling candidate phrases determined by each $s_i$ and choosing phrases using beam search \cite{SutskeverVL14}. As the generation procedure is at the phrase-level, we evaluate each generated question by comparing the generated phrases with the original and averaging evaluation results over all the phrases in the questions. 

We adopt the following three standard metrics to measure the quality of the generated questions from lexical, semantic and diversity perspectives. 

\begin{itemize}
    \item  \emph{Smoothed Sentence-level} BLEU~\cite{PapineniRWZ02,ChenC14}: BLEU is a popular metric to measure the geometric mean of modified $n$-gram precision with a length penalty. As $N$ new questions are generated, we define the $n$-gram precision and $n$-gram recall as the average and the maximum value of $N$ $n$-gram BLEU scores respectively. We use $3$-gram with smoothing technique, and BLEU scores are normalized to $[0, 1]$.
    \item  \emph{Cosine similarity of Bag-of-words (BOW) embeddings}: a metric matches phrase embeddings through the average, extreme or greedy strategy over all the word embeddings in the phrases \cite{forgues2014bootstrapping,RusL12}. The score is the cosine distance between the two produced vectors. We used pretrained embeddings \footnote{Implementation details are in supplementary material. \emph{Average}: cosine similarity between the averaged word embeddings; \emph{Extrema} \cite{forgues2014bootstrapping}: cosine similarity between the biggest extreme values among the word embeddings of the two phrases; \emph{Greedy} \cite{RusL12}: matching words in two phrases greedily based on their embeddings' cosine similarity and averaging the obtained scores.} and denote the three metrics as ``Average'', ``Extreme'' and ``Greedy''.
    \item \emph{Distinct} \cite{DialogWAE}: a metric  computes the diversity of the generated phrases. The ratio of unique $n$-grams over all $n$-grams in the generated phrases is denoted as distinct-$n$. We further define \emph{intra-dist} as the average of distinct values within each sampled phrase and \emph{inter-dist} as the distinct value among all sampled phrases.
\end{itemize}

We compare the proposed method eg-CVAE with the aforementioned baselines on the collected real-word NMLEC\_QA dataset, and report the experiment results in Table \ref{auto_result}. The highest score in each column is in bold for clarity. In the following, we discuss the results in details.

First, we examine the results in terms of similarity using BLEU and BOW metrics. Our proposed method eg-CVAE is designed to promote diversity, and thus the semantic similarity score is not that high. The vanilla CVAE-based VHRED does not involve any constraint on the latent distribution of $z$, and the HRED \cite{SerbanSBCP16} models the decoding process in a definite way without further manipulation on the hidden context, so their semantic similarity scores are medium. A variant of the proposed method type-CVAE models prior type information, and another variant entity-CVAE models entity explicitly. These constraints facilitate models to generate more similar QA pairs to the original.

On the other hand, from the view of diversity, the proposed method eg-CVAE has the highest score over distinct metrics. This is because that we hierarchically generate new questions based on the latent answer-conditioned distribution, rather than a definite decoding process. As pointed out in \cite{serban2017hierarchical}, this hierarchical strategy can prevent diversity being injected at the low level.

\subsection{Human Evaluations \protect \footnote{We also propose a reusable method for evaluation using human annotation of key phrases in supplementary material.}}
Following \cite{li2018improving}, we further conduct human evaluation on 10\% samples from NMLEC\_QA training dataset and the corresponding generated QA pairs by our methods and baselines. Three experts (real doctors) were asked to assess each QA pair from three perspectives: 1) Consistent: How consistent the generated QA is compared with the original one? 2) Informative: How informative the generated QA is against the original one? 3) Fluent: How fluent the phrases of a generated question are? Each perspective is assessed with a score from 1 (worst) to 5 (best). The average results are presented in Table \ref{my_human}. 

\begin{table}
\centering
 \caption{Human evaluation results.$^{*}$ indicates the difference between eg-CVAE and other baselines are statistically significant $(p < 0.01)$ by two-tailed t-test.}
\label{my_human}
\begin{tabular}{cccc}
\toprule
\textbf{Method} & \textbf{Consist.} & \textbf{Informat.} & \textbf{Fluency} \\
\midrule
HRED & 3.68$^{*}$ & 3.38$^{*}$  & 3.93$^{*}$ \\
VHRED & 2.79$^{*}$ & 3.52 & 3.79$^{*}$ \\
type-CVAE & 3.53$^{*}$ & 3.42 & 4.03$^{*}$ \\
entity-CVAE & 3.68$^{*}$ & 3.38$^{*}$ & 4.08$^{*}$ \\
eg-CVAE & \textbf{4.09} & \textbf{3.62}  & \textbf{4.43} \\
\bottomrule
\end{tabular}
\end{table}

The results show that our model consistently outperforms the seq2seq-baseline model (HRED) and the vanilla CVAE-based method (VHRED). 
The type-level and entity-level modelings of medical questions make the key information consistent. The prior information from these two levels of modeling also ensures the good ability of our model to generate informative and fluent questions. 

Moreover, the implicit type-level modeling via aggregated embedding introduces more variance but less consistence against explicit entity-level modeling via concrete entities, which inspires us to combine them together in the eg-CVAE.

\subsection{Qualitative Analysis}
To further qualitatively analyze the proposed method through real cases, we  compare the generated QA pairs from different models in Figure \ref{tb:case1}. Each example consists of an original valid QA pair and three generated questions, which are sampled based on the raw one through beam-search. We can clearly see our eg-CVAE retains both one-to-many diversity property and validity of each phrase's generation.

We compared three models here including HRED, CVAE and eg-CVAE.\footnote{We include detailed case comparison between eg-CVAE, type-CVAE and entity-CVAE in supplementary material.} For HRED, we can observe that the generated questions' diversity is limited since the model tends to repeat the seed phrases (e.g., the meaningless repetition of ``RBC'' and ``anxiety'') and the important information describing topographic shape (e.g., ``lower than'' in ``HB is lower than normal'') is lost. On the contrary, CVAE explores the discourse-level diversity but ambiguous phrases like ``wbc $3.45\times10^{12}$/l'' in $Q1$, which indicates potential inflammation rather than anemia, are often generated in a key place. Similarly, in $Q3$ from CVAE  ``sudden fever after menstruation, discomfort'' in most cases indicates endocrine disorders rather than anemia.

For eg-CVAE, we can see it explores discourse-level diversity by generating symptoms like ``whitish complexion'' in $Q1$ that are not existing in the $Q$. In terms of the validity, the generated imperative semantics of the non-key phrases are consistent with the implicit semantics of the original questions of anemia. For example, although the semantics of ``the poor face'', ``anxious'' and ``whitish complexion'' in $Q$ and $Q3,Q1$ are different, they does not influence on the overall diagnosis of ``anemia''. The generated ``the normal systolic blood pressure'' and ``normal liver'' do not affect the judgment of ``anemia'' as they are normal body signal, too.

\subsection{Evaluation on a QA System}
To further study the usefulness of the generated medical QA pairs, we integrate such generated pairs into a QA system, which is an attention-based model \cite{cui2017attention} for NMLEC\_QA dataset. The results are summarized in Table \ref{qa_result}. 

\begin{table}
\centering
\caption{Usefulness of the generated QA pairs. $^{*}$ indicates difference between the original setting and the new setting is statistically significant ($p < 0.01$). \protect \footnotemark[7]} 
\label{qa_result}
\begin{tabular}{cc}
\toprule
\textbf{Dataset} & \textbf{Accuracy} \\
\midrule
Original & 61.97 \\ 
+ HRED QA & 58.78 \\ 
+ VHRED QA  & 62.28 \\ 
+ type-CVAE QA & 65.27$^{*}$ \\ 
+ entity-CVAE QA & 64.67$^{*}$ \\ 
+ eg-CVAE QA & \textbf{67.96}$^{*}$ \\ 
\bottomrule
\end{tabular}
\end{table}
\footnotetext[7]{We calculate statistical significance based on the bootstrap test in \cite{noreen1989computer} with 10k samples.}

\begin{figure*}
\begin{center}
\includegraphics[width=0.8\textwidth]{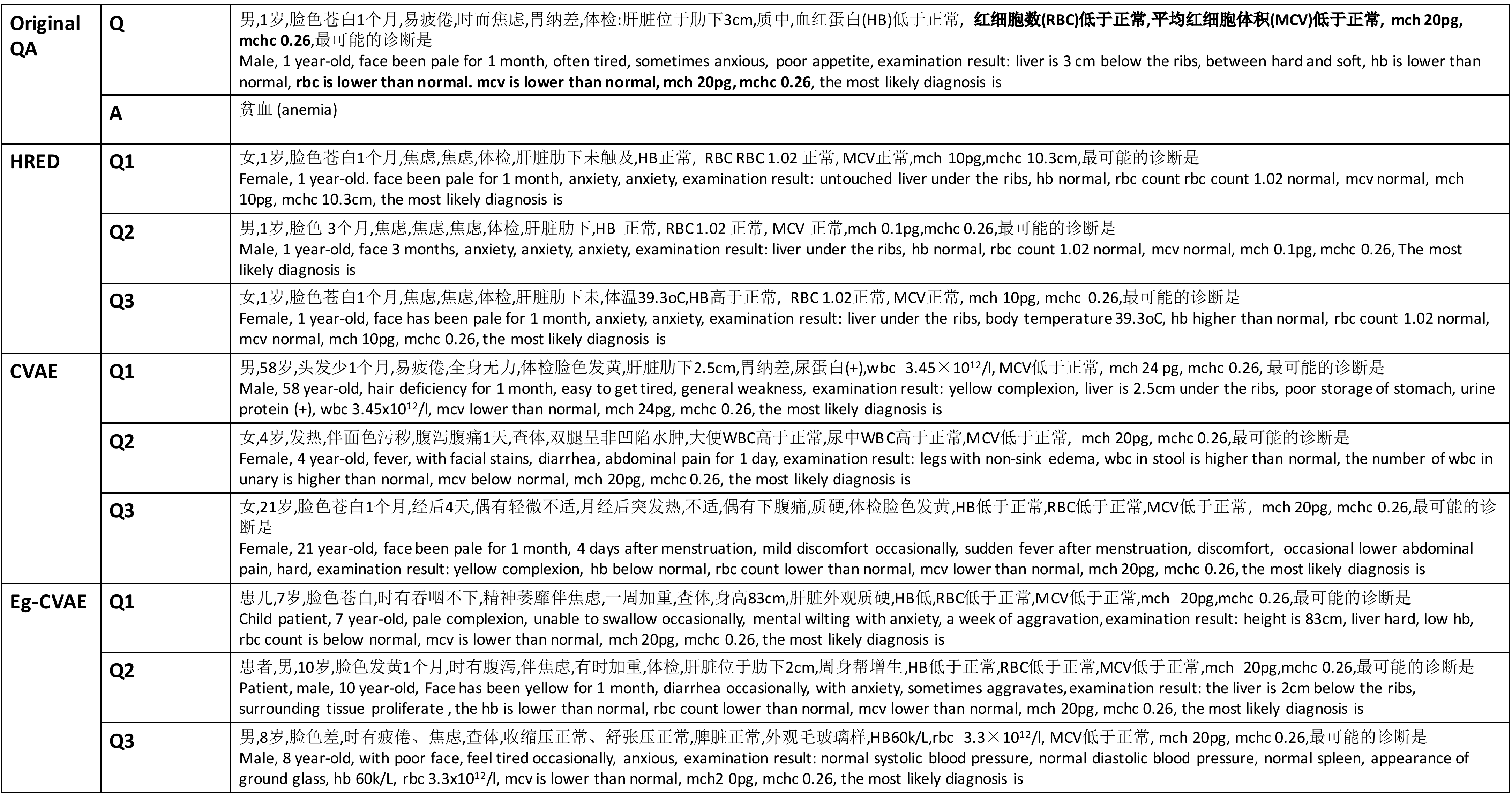}
\caption{Case study for generated QA pairs of different methods (the key phrases in original QA pair are in bold)}\label{tb:case1}
\end{center}
\end{figure*}

For baseline methods, integrating the generated QA pairs from HRED hurts the accuracy without augmented data. As pointed out in \cite{serban2017hierarchical}, HRED is very likely to favor short-term predictions instead of long-term predictions. As shown in Figure \ref{tb:case1}, rather than globally considering context phrases to generate a meaningful phrase for the current slot, HRED tends to repeat the predicted correct word. The lack of diversity and repeat of common words lead to the discrepancy in the generated questions' distribution and the original one, which may cause the degradation and introduce noise to the original dataset. From the results of vanilla CVAE-based VHRED, we can see that the improvement exits but is marginal. we presumes that is because the lack of constraint on the latent distribution leads to weak guidance from the corresponding answer and the unlabeled textbook for generated questions from VHRED.

Two variants of the proposed method, entity-CVAE and type-CVAE, generated QA pairs that boost the original QA system with 3-4\% accuracy. Each of them introduces external constraints on the latent variable in the decoding phase, which may help to diversify the generated questions while keeping linguistic and structural relationships within original questions. Furthermore, type-CVAE generates QA pairs that seem to be more helpful to the QA system. This benefit may come from the modeling of type information, which allows the generated questions to be relatively more diverse and thus introduces more useful knowledge. The proposed method eg-CVAE combines the advantages of entity-CVAE and type-CVAE, building a three-pass decoding process, and thus improves the QA system to achieve the highest accuracy. These observations further demonstrate the usefulness of the generated medical QA pairs by eg-CVAE.

\section{Related Work}
Question Generation \cite{heilman2010good} has attracted increasing attention in recent years. However, most existing work only focuses on the similarity of generated questions with the original ones, but ignores the usefulness in training a QA system of generated questions given answers. Earlier work in question generation employed rule-based approaches to transform input texts into corresponding questions, usually requiring some well-designed general rules \cite{mitkov2003computer}, templates \cite{labutov2015deep} or syntactic transformation heuristics \cite{ali2010automation}. Recent studies leveraged neural networks to generate questions in an end-to-end fashion. \cite{DuSC17} applied the attention-based sequence-to-sequence model to generate questions in the context of reading comprehension. In medical QA, \cite{roberts2017overview,pampari2018emrqa} targets the same problem as us from the dataset angle. \cite{walonoski2017synthea} is similar to us, but they focus on the state transition of patient records.

Other existing work, which tackles the usefulness and models the question-answer pair generation directly, still sets the diversity of questions for the corresponding answer aside and requires related context in prior. \cite{serban2016generating} applied the encoder-decoder framework to generate question-answering pairs from built knowledge base triples. \cite{subramanian2018neural} formulated the question-answer pair generation in reading comprehension, where each pair will be given one high-quality context and the answer is a text span of the context, separately with the answer detection and question generation problem. \cite{YuanWGSBZST17} leveraged policy gradient techniques to further improve the generation quality. Coreference knowledge is also introduced for question-answer pair generation from Wikipedia articles with the context in \cite{CardieD18}.  \cite{duan2017question} investigated integrating generated questions from given context to the question-answering system on sentence selection tasks, which leveraged both rule-based features and neural networks to approximate the semantics of generated questions with original ones. \cite{yang2017semi,song2018leveraging} also leveraged the generated QA for QA system. But they all have the external context in SQuAD~\cite{rajpurkar2016squad} to build upon, which does not exist in our medical setting.

Compared to existing work, our work introduces structure information of QA pairs generation in medical domain, which does not involve any prior context. To ensure the validity of generated QA pairs, we proposed an unsupervised detector to automatically explore external materials. We also proposed to model the question-answer pair generation problem directly as approximating the latent distribution of medical questions with the corresponding answer. 

\section{Conclusions}
In this paper, we introduced a novel framework, consisting of an unsupervised key phrase detector and an Entity-guided CVAE-based generator, for automated question-answer pair generation in the medical domain. 
Different from existing seq2seq models that involve a  definite encoding-decoding procedure to restrict the generation scope, or traditional CVAE models that directly approximate the posterior distribution over the latent variables to a simple prior, the proposed method models the generation process as a multi-pass procedure (type, entity and phrase as constraints over the latent distribution) to ensure both validity and diversity. 
Experiments on a real-world dataset from the National Medical Licensing Examination of China demonstrate that the proposed method outperforms existing methods and can generate more diverse, informative and valid medical QA pairs that further benefit the examination QA system. We will investigate more on the generalizability of proposed method on standard dataset like SQuAD~\cite{rajpurkar2016squad} and its integration with popular pretrained model~\cite{devlin2018bert} in the future work.

\section{Supplementary Material}
\subsection{Implementation Details}
The proposed method is trained with the following hyperparameters: Word embedding is pre-trained using the whole unstructured medical materials with a vector dimension of $200$, and the learned vector representations are shared across different components of the proposed method. 
The phrase encoder's hidden dimension is set to be $300$. 
The hierarchical context phrase encoder has a hidden dimension of $600$, and the latent variable $z$ has a size of $200$. The number of retrieved medical text is set to be $10$. The size of sliding window in hierarchical pooling method is set to $3$.
Both the prior network and the MLP for one-pass type decoder have one hidden layer of size $400$ and tanh non-linearity activation function. The two-pass entity decoder is another MLP with the dimension of the entity vocabulary size. By connecting to a softmax layer, an entity embedding with a dimension of $50$ is then applied for aggregation. The final phrase decoder's hidden dimension is set to be $400$. The initial weights of these networks are sampled from a uniform distribution $[-0.08, 0.08]$. 
The models are trained with a mini-batch size of $30$, Adam optimizer with a learning rate of $0.001$, and gradient clipping at $5$. Further, we use the BOW loss along with KL annealing of $10,000$ batches. We conduct these parameter selections based on the variational lower bound.

\subsection{Human Evaluation}
\begin{figure}
\begin{center}
\includegraphics[width=0.85\columnwidth]{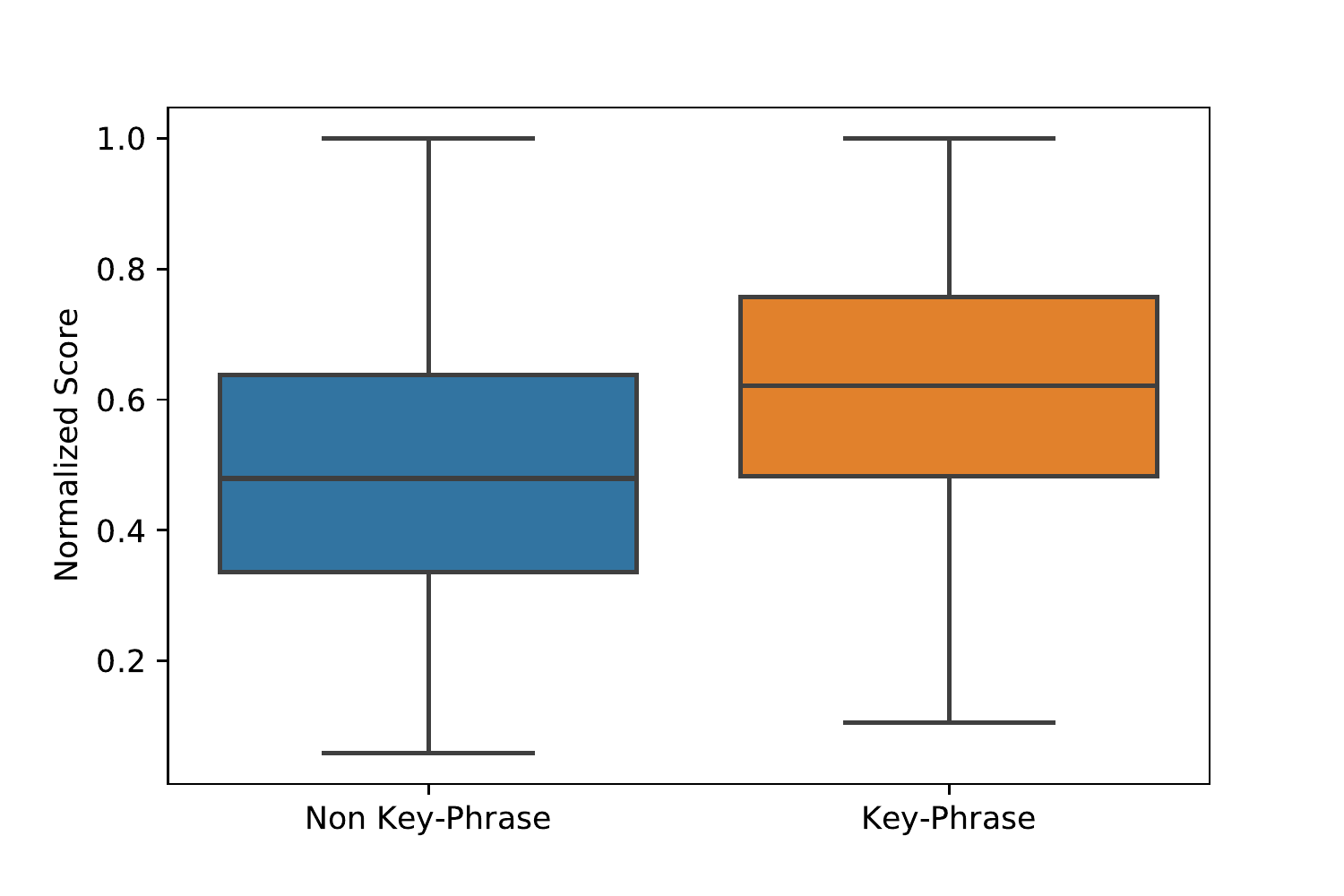}
\caption{The distribution of proposed score for key phrases.}\label{fig:quantitive_score}
\end{center}
\end{figure}

We also propose a reusable method for evaluation using human annotation of key phrases. As mentioned above, we assume the presence of these highly answer related key phrases in a generated question indicates it is likely to match the corresponding answer. We thus employ three experts to label the key phrases in $500$ random sampled medical QA pairs from NMLEC\_QA dataset. The labeled key phrases with at least two experts' consensus will have the final label of ``Yes", others ``No". To evaluate the unsupervised key phrase detector in our proposed method eg-CVAE, we plot the distribution of the key phrase score this detector assigns to all labeled data in Figure \ref{fig:quantitive_score}. From this plot, we can see that the detector could assign the real key phrases with higher scores, which ensures the higher probability of these key phrases to be unchanged and facilitates our model to generate medical questions that match the conditioned answers.

\subsection{Case Study}
To qualitatively analyze the proposed method through some real cases, we first compare the generated QA pairs from different models in Figure \ref{tb:case1}. Secondly, we present a detailed illustration of our proposed method entity-guided CVAE's generation process in Figure \ref{tb:case2}. 

\subsubsection{Case study for generated QA pairs from different models.}
We demonstrate several generated medical QA pairs from different models in Figure \ref{tb:case1}. Each example is consisted of an original valid QA pair and three generated questions, which are sampled based on the raw one through beam-search.  Three models are compared with including HRED, CVAE and eg-CVAE.

\begin{enumerate}
\item For HRED, it is easy to find that the generated questions' diversity is limited since the model tends to repeat the seed phrases (e.g., the meaningless repetition of ``RBC'' and ``anxiety'') and the important information describing topographic shape (e.g., ``lower than'' in ``HB is lower than normal'') is easily lost. Since the model does not distinguish the phrases between each other, the lexical metric (BLEU score) which measures $n$-gram exact match with the original question seems to be high, even though the generated question and the corresponding answer do not match.
\item For CVAE, it is obvious that CVAE explores the diversity in generation. However, the sampled questions show that ambiguous phrases are often generated in a key place. For instance, in the first sentence ``wbc $3.45\times10^{12}$/l'' is very likely to indicate inflammation, in the second sentence excessive symptoms of ``diarrhea'' may guide to either anemia or diarrhea, and in the third sentence  ``sudden fever after menstruation, discomfort'' in most cases indicates endocrine disorders rather than anemia. This is due to two reasons - the model involves limited constraints on the latent distribution from the answer and the iterative generation setting makes the model focus more on the previous generated phrases, which to some extent weakens the restriction from answer on the whole generation.
\item For eg-CVAE, we can clearly see that this model retains the one-to-many diversity property of each phrase's generation. Moreover considering the validity, the generated imperative semantics of the non-key phrases are consistent with the implicit semantics of the original questions of anemia. For example, although the semantics of ``the poor face'', ``anxious'' and ``whitish complexion'' are different, they does not influence on the overall diagnosis of ``anemia''.  For more examples, ``the normal systolic blood pressure'' and ``normal liver'' have no influence on the judgment of ``anemia'' as they are both normal body sign.
\end{enumerate}

\subsubsection{Detailed case study for generation process of  entity-guided CVAE.}
\begin{figure*}
\begin{center}
\includegraphics[width=0.99\textwidth]{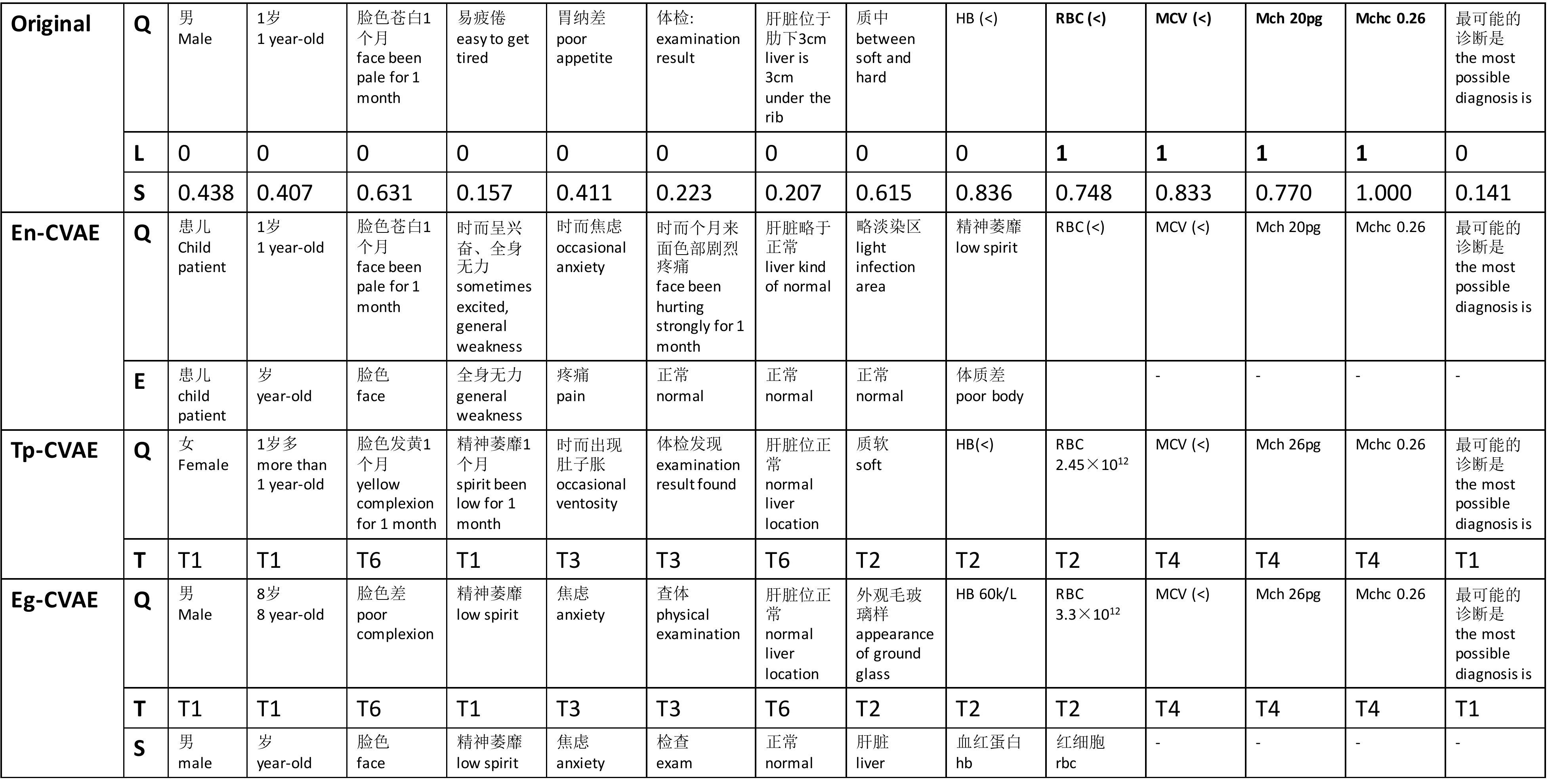}
\caption{Further case study for generated QA pairs of eg-CVAE. (Q, L, S, T and E stand for question, label, score, type and entity,  respectively. $(<)$ indicates that ``lower than normal'')}\label{tb:case2}
\end{center}
\end{figure*}

To further demonstrate the advantages of the proposed eg-CVAE in terms of diversity and validity in the above analysis, we present in detail the generation process involving different constraints with respect to the latent variable $z$ (multiple-pass decoding procedure) in Figure \ref{tb:case2}.

From the results, on the one hand, we can observe that explicit entity modeling at first-pass makes the generated phrases strongly related to the modeled entities. Many en-CVAE generated phrases directly contain the modeled entities, and the diversity is relatively limited. Moreover, once the decoded entities are relatively abstractive, (e.g., ``poor body''), the generated phrase may not contain the key information in the original question, such as informative phrase ``HB is lower than normal'' replaced by trivial phrase ``low spirit''. On the other hand, implicit type modeling at first-pass encourages more diversity in generation. Since the constraint extent on type by decoding the contextualized type vector is much looser than model with the decoding explicit entities, the generated diversity will be more broad, such as ``occasional ventosity'' or ``yellow complexion'', etc.

To handle this phenomenon, eg-CVAE comprehensively treats explicit entity modeling and implicit type modeling as different decoding passes. By modeling type information and then introducing it as a priori to the entity modeling, eg-CVAE prevents the loss of key information; and by adding variants through multiple-pass decoding processes, generated questions are well diversified. In this way, the diversity and validity of generated QA pairs are both guaranteed.

\small
\bibliographystyle{aaai}
\bibliography{AAAI2888}

\end{document}